\documentclass[sigconf]{aamas}  

\usepackage{booktabs}
\usepackage{amsmath,amssymb,amsfonts}
\usepackage{algorithm}
\usepackage{algorithmic}
\usepackage{graphicx}
\usepackage{subcaption}
\usepackage{textcomp}
\usepackage{xcolor}
\usepackage{epsfig}
\usepackage[symbol]{footmisc}

\setcopyright{ifaamas}  
\acmDOI{doi}  
\acmISBN{}  
\acmConference[AAMAS'18]{Proc.\@ of the 17th International Conference on Autonomous Agents and Multiagent Systems (AAMAS 2018), M.~Dastani, G.~Sukthankar, E.~Andre, S.~Koenig (eds.)}{July 2018}{Stockholm, Sweden}  
\acmYear{2018}  
\copyrightyear{2018}  
\acmPrice{}  

\newcommand{\figref}[1]{Fig.~\ref{fig:#1}}
\newcommand{\algoref}[1]{Algorithm~\ref{algo:#1}}

\newcommand{\tabref}[1]{Table~\ref{tab:#1}}
\newcommand{\eqreff}[1]{Eq.~\ref{eq:#1}}

\newcommand{\figlabel}[1]{\label{fig:#1}}
\newcommand{\algolabel}[1]{\label{algo:#1}}

\newcommand{\tablabel}[1]{\label{tab:#1}}
\newcommand{\eqlabel}[1]{\label{eq:#1}}

\begin{document}

\title{Inverse Kinematics and Sensitivity Minimization of an n-Stack Stewart Platform}  

\subtitle{Robotics Track}




\author{David Balaban}
\authornote{former intern at NASA Langley Research Center, Structural Mechanics and Concepts Branch}
\affiliation{
	\institution{University of Massachusetts Amherst \\ College of Computer Sciences}
	\city{Amherst}
	\state{Massachusetts}
	\postcode{01003}
}
\email{dbalaban@cs.umass.edu}

\author{John Cooper}
\affiliation{
	\institution{NASA Langley Research Center \\ Dynamic Systems and Control Branch}
	\city{Hampton}
	\state{Virginia}
	\postcode{23666}
}
\email{john.r.cooper@nasa.gov}

\author{Erik Komendera}
\authornote{formerly NASA Langley Research Center, Structural Mechanics and Concepts Branch}
\affiliation{
	\institution{Virginia Tech \\ Dept. of Mechanical Engineering}
	\city{Blacksburg}
	\state{Virginia}
	\postcode{24061}
}
\email{komendera@vt.edu}

\begin{abstract}  
An autonomous system is presented to solve the problem of in space assembly, which can be used to further the NASA goal of deep space exploration. Of particular interest is the assembly of large truss structures, which requires precise and dexterous movement in a changing environment. A prototype of an autonomous manipulator called "Assemblers" was fabricated from an aggregation of Stewart Platform robots for the purpose of researching autonomous in space assembly capabilities. The forward kinematics for an Assembler is described by the set of translations and rotation angles for each component Stewart Platform, from which the position and orientation of the end effector are simple to calculate. However, selecting inverse kinematic poses, defined by the translations and rotation angles, for the Assembler requires coordination between each Stewart Platform and is an underconstrained non-linear optimization problem. For assembly tasks, it is ideal that the pose selected has the least sensitivity to disturbances possible. A method of sensitivity reduction is proposed by minimizing the Frobenius Norm (FN) of the Jacobian for the forward kinematics. The effectiveness of the FN method will be demonstrated through a Monte Carlo simulation method to model random motion internal to the structure.
\end{abstract}



\maketitle


\section{Introduction}
NASA is tasked with developing technologies for deep space exploration and habitation~\cite{howe2013nasa}\cite{kennedy2011nasa}. To further that goal, NASA is developing a robotic assembly process of deep space structures~\cite{cohen1997habitats}\cite{rojas2012analysis}\cite{doggett2002robotic}. A recent robotics concept introduces the use of coordinating Stewart platforms~\cite{bingul2012dynamic} arranged in a stack, called Assemblers. Assemblers may come in different size stacks of at least one Stewart platform. \figref{assembler} shows a picture of a prototype made of four Stewart platforms. The Assemblers are intended to be used in coordination with other such robots, to arrange themselves into multiple potential topologies, including the ability to self-assemble from smaller stacks of Stewart Platforms.

These robots have a complicated geometric structure with nonlinear constraints and many degrees of freedom (DOF) internal to the structure, leading to over-actuated forward kinematics in stacks greater than one platform. An over-actuated system has more actuators than the end effector has DOF. The method of Frobenius norm (FN) minimization is presented~\cite{custodio2010incorporating} to select the inverse kinematic pose which optimizes structural sensitivity.

\begin{figure}
	\begin{center}
		\includegraphics[width=.5\linewidth]{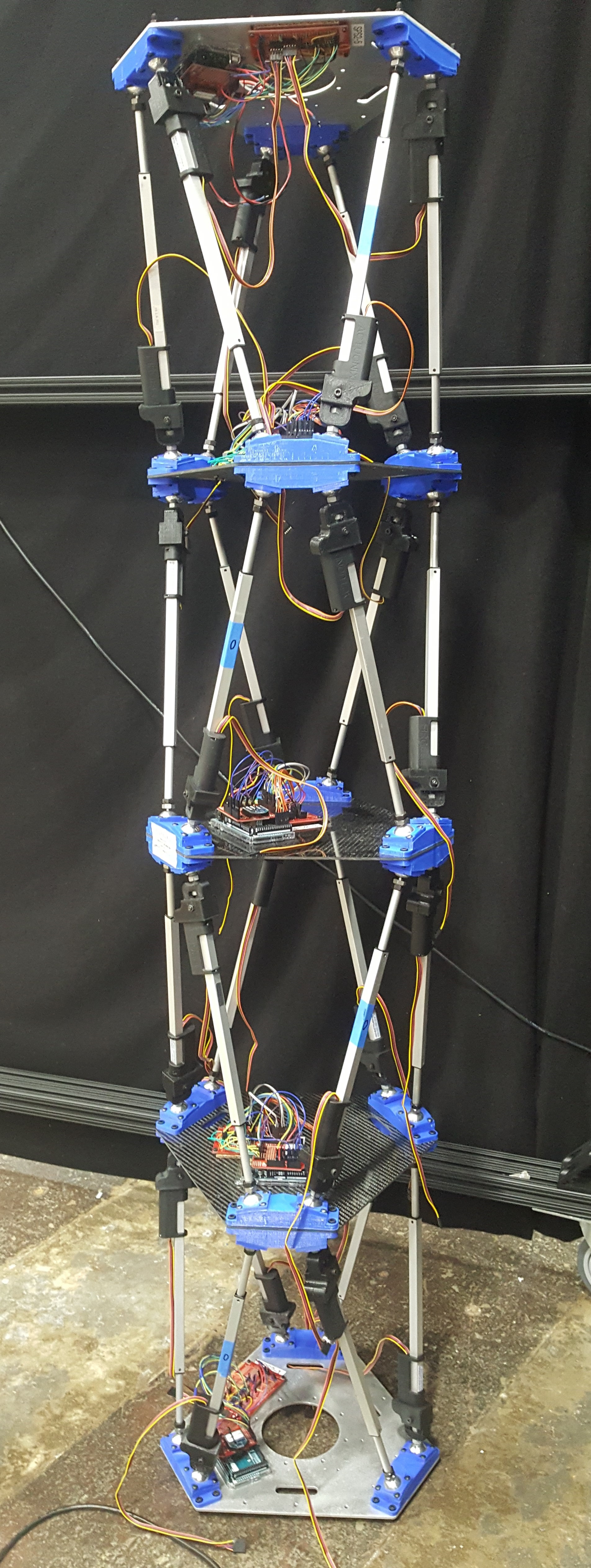}
	\end{center}
	\caption{Picture of a 4-stack Assembler prototype}
	\figlabel{assembler}
\end{figure}

From the 1970s and into the 1990s, NASA researched the construction of structures in low Earth orbit during the servicing and construction of the Hubble Space Telescope and the International Space Station. These projects relied on astronauts to service them~\cite{watson2002history}.

Past research on in-space assembly (ISA) have been focussed on reliable deployments and assembly of truss structures~\cite{watson2002history}. Truss structures are useful for space exploration because they can be unpackaged after launch and provide the structural stiffness and maneuverability necessary for operations in-space~\cite{rhodes1985deployable}\cite{mikulas1987deployable}.

In order to support deep space exploration activities, where manned assembly of large structures is impractical, NASA is developing autonomous construction and ISA methods that replace humans in dangerous environments~\cite{zimpfer2005autonomous}. Much of this research is focused on the robotic assembly of truss structures~\cite{doggett2002robotic}.

One concept involves an autonomous manipulator known as Intelligent Precision Jigging Robot (IPJR) used for precision welding~\cite{komendera2014truss}. Several other proposals exist for in-orbit assembly such as the Commercial Infrastructure for Robotic Assembly and Servicing (CIRAS), SpiderFab, Archinaut and Dragonfly projects~\cite{roarobotic}. The Assemblers are proposed to add mobility and adaptability to enable assembly on extra-terrestrial surfaces such as the moon or Mars.

Robotic ISA requires precise manipulation in a changing extraterrestrial environment. The Assemblers were designed as a modular robot, which provides several adaptability benefits for working in such spaces~\cite{yim2002modular}\cite{miura1988adaptive}. While the forward and inverse kinematic properties of the component Stewart platform structures are well known~\cite{abdellatif2009computational}, there has been limited work on stacks of Stewart platforms~\cite{williams1994kinematic}\cite{hamlin1996tetrobot}.

The contributions of this paper include the formulation of a solver which can efficiently find inverse kinematic solutions for an Assembler with an arbitrary number of platforms, and the demonstration of optimal pose selection via FN minimization with numerical results.

\section{Stewart Platforms}
Stewart platforms, or parallel plate manipulators, consist of two plates adjoined by six linear actuators. \figref{platform} shows a simplified diagram of a Stewart Platform with the coordinates we use. To describe the inverse kinematics of the Stewart platform, we consider two reference frames attached to the top and bottom plates. The plane of the top plate is defined by the perpendicular normal vectors $\hat{t}_x$ and $\hat{t}_y$, while the bottom plate is defined by $\hat{b}_x$ and $\hat{b}_y$. The unit normal vector to the bottom and top plates are given by $\hat{n}_b = \hat{b}_x\times\hat{b}_y$ and $\hat{n}_t = \hat{t}_x\times\hat{t}_y$ respectively. The orthonormal basis sets defining the coordinate space for each plate are therefore given by $[\hat{b}_x, \hat{b}_y, \hat{n}_b]$ and $[\hat{t}_x, \hat{t}_y, \hat{n}_t]$, and the origin of each reference frame is the center of the respective plate. Let $\mathbf{R}$ be a rotation matrix which brings a vector from the reference frame of the top plate into that of the bottom plate. In the reference frame of the bottom plate, let $\vec{P}$ be the vector which points from the center of the bottom plate to the center of the top plate.

In \figref{platform} each leg is adjoined to the plates at nodes, which represent ball and socket joints. Each actuator connects a single node on each plate. Let $\vec{t}$ describe a node position in the top plate, and let $\vec{b}$ describe likewise for a bottom plate node. Vectors $\vec{t}$ and $\vec{b}$ point from the center of their respective plate to the relevant node in the respective reference frames. The actuator vector in the bottom plate's reference frame is therefore given by:

\begin{align}
\vec{\ell} = \mathbf{R} \vec{t} + \vec{P} - \vec{b}
\end{align}

\begin{figure}
	\begin{center}
		\includegraphics[width=.7\linewidth]{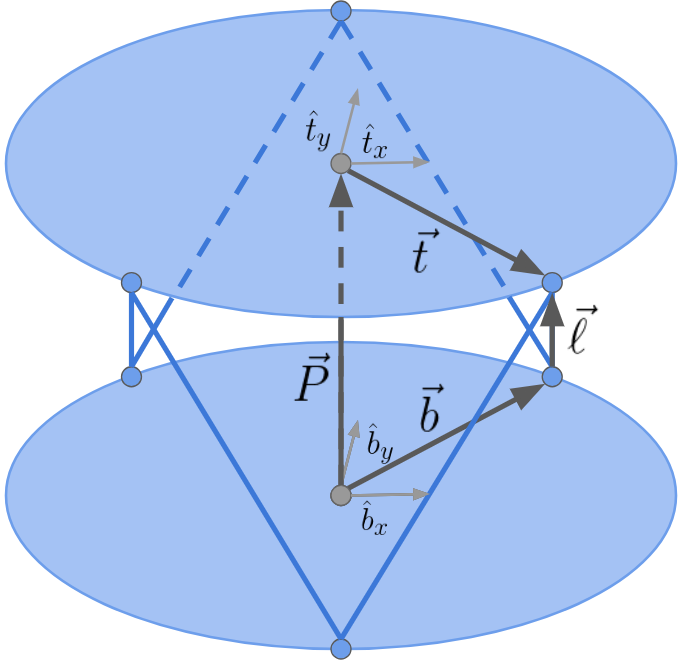}
	\end{center}
	\caption{Diagram of Stewart Platform Coordinates}
	\figlabel{platform}
\end{figure}

This equation describes the inverse kinematics for a single Stewart Platform: given a desired translation vector $\vec{P}$ and orientation $\mathbf{R}$, the lengths of each actuator should be set to $||\vec{\ell}||$. The forward kinematics are more difficult to compute, and require a non linear solver supplied with an initial guess~\cite{jakobovic2002forward}.

Each Stewart Structure has geometric constraints which limit the set of reachable poses. There are the length constraints on the extension and retraction of the actuators, as well as angular constraints set by the ball joint design. These constraints are given by:

\begin{align} \eqlabel{constraints}
\ell_{min}^2 \le \vec{\ell}\ ^T \ \vec{\ell} \le \ell_{max}^2 \\
\vec{\ell}\ ^T \hat{n}  \ge ||\vec{\ell}|| \sin(\theta_{min}) \\
\vec{\ell}\ ^T \mathbf{R} \hat{n} \ge ||\vec{\ell}|| \sin(\theta_{min})
\end{align}

Where $\hat{n} = \begin{bmatrix} 0 & 0 & 1 \end{bmatrix}^T$ is the vector normal to the top plate in its own reference frame, $\theta_{min}$ is the minimum angle allowed by the ball joint, and $\ell_{max}$, $\ell_{min}$ are the maximum and minimum possible actuator lengths respectively. The square of the length is used to avoid the use of the square root operation to simplify derivatives.

\section{n-Stack Assembler}

An Assembler is composed of $n$ Stewart Platforms, where the top plate of one platform is directly adjoined to the bottom plate of the one above it. The topmost plate is the end effector, while the bottommost plate is the baseplate. 

The pose of an Assembler is defined by the set of translations and rotations of each composite platform: $\mathbf{R}_i, \vec{P}_i$ for $i = 1 \ldots n$. The baseplate's position and orientation defines the global reference frame, and the end effector global position and orientation are $\vec{P}_{ee}$ and $\mathbf{R}_{ee}$ respectively. An Assembler is similar to a serial manipulator with variable link lengths. In this domain, the forward kinematics can be expressed as:

\begin{align} \eqlabel{forward}
\vec{P}_{ee} &= \vec{P}_1 + \sum_{i=2}^{n}\prod_{j=1}^{i-1}(\mathbf{R}_j)\vec{P}_i \\
\mathbf{R}_{ee} &= \prod_{i=n}^{1}\mathbf{R}_i
\end{align}

Each platform has 6 DOF, which means an Assembler has $6n$ DOF. Therefore, the end effector, which has 3 translational DOF and 3 rotational DOF, is underconstrained. Ideally, an autonomous Assembler should choose the pose which has the least sensitivity to disturbances when confronted with many feasible solutions to enhance precision in the end effector. The inverse kinematic problem for Assemblers is therefore to choose all $\mathbf{R}_i$ and $\vec{P}_i$ such that the desired end effector pose is reached, without violating the geometric constraints, and while giving the structure as little sensitivity to internal movement as possible.

\section{Structural Sensitivity}

To formulate a solver which minimizes structural sensitivity, it is necessary to quantify that property from the forward kinematics. We compare structural sensitivity using the FN of the end effector Jacobian.

We can model movement in the end effector as deriving from small changes to the internal structure of the Assembler. Let $\mathbf{x}$ be the Assembler state vector containing x-y-z translation and yaw-pitch-roll angle quantities from all $\mathbf{R}_i, \vec{P}_i$. Let $\mathbf{y}$ be the end effector output vector containing translation and orientation quantities in $\vec{P}_{ee}$ and $\mathbf{R}_{ee}$. Let $f(\cdot)$ describe the forward kinematics such that $\mathbf{y} = f(\mathbf{x})$, and $\mathbf{J_{ee}} = \frac{df}{dx}$ be the end effector Jacobian. For a small change $\mathbf{\delta x}$ in the state vector, the resulting change in the end effector is given by $\mathbf{\delta y}$. To approximate the non-linear relationship between $\mathbf{\delta x}$ and $\mathbf{\delta y}$, we take the first order term of the Taylor Series expansion of $f(\cdot)$:

\begin{align}
\mathbf{\delta y} = \mathbf{J_{ee} \delta x}
\end{align}

The FN, defined as $||\mathbf{J_{ee}}||_F = \sqrt{Tr(\mathbf{J_{ee}^T J_{ee}})}$, has the property:

\begin{align}
||\mathbf{\delta y}|| \le ||\mathbf{J_{ee}}||_F ||\mathbf{\delta x}||
\end{align}

Therefore, the FN of the end effector Jacobian puts an upper bound on how much the end effector can be affected by internal motion. By minimizing the FN we lower that bound for enhanced sensitivity in the end effector, which comes solely from the geometry of the structure.

The end effector orientation is defined by the matrix $\mathbf{R_{ee}}$, this quantity can either be reduced to a vector by extracting a set of Euler angles, or by elongating the matrix into a vector of size 9. Let $\vec{R}_{ee}$ be the elongated form of $\mathbf{R_{ee}}$. Note that for any matrix $\mathbf{M}$ with elongated form $\vec{M}$, the following relationship holds: Trace$(\mathbf{M^T M}) = \vec{M} \cdot \vec{M}$. The Jacobian is therefore written as:

\begin{align}
&\mathbf{J_{ee}} = \begin{bmatrix}
\frac{\partial \vec{P}_{ee}}{\partial \vec{P}_i} & \frac{\partial \vec{P}_{ee}}{\partial \theta_{ij}} \\ \frac{\partial \vec{R}_{ee}}{\partial \vec{P}_i}
& \frac{\partial \vec{R}_{ee}}{\partial \theta_{i j}}
\end{bmatrix} \\
&\mathbf{J_{ee}^TJ_{ee}} = \\ \notag &\begin{bmatrix} 
\frac{\partial \vec{P}_{ee}}{\partial \vec{P}_i}^T \frac{\partial \vec{P}_{ee}}{\partial \vec{P}_i} + \frac{\partial \vec{P}_{ee}}{\partial \theta_{ij}}^T \frac{\partial \vec{P}_{ee}}{\partial \theta_{ij}} & 
\frac{\partial \vec{R}_{ee}}{\partial \vec{P}_i}^T \frac{\partial \vec{P}_{ee}}{\partial \vec{P}_i} + \frac{\partial \vec{R}_{ee}}{\partial \theta_{i j}}^T \frac{\partial \vec{P}_{ee}}{\partial \theta_{ij}} \\ \frac{\partial \vec{P}_{ee}}{\partial \vec{P}_i}^T\frac{\partial \vec{R}_{ee}}{\partial \vec{P}_i} + \frac{\partial \vec{P}_{ee}}{\partial \theta_{ij}}^T\frac{\partial \vec{R}_{ee}}{\partial \theta_{i j}} & \frac{\partial \vec{R}_{ee}}{\partial \vec{P}_i}^T\frac{\partial \vec{R}_{ee}}{\partial \vec{P}_i} + \frac{\partial \vec{R}_{ee}}{\partial \theta_{i j}}^T\frac{\partial \vec{R}_{ee}}{\partial \theta_{i j}}
\end{bmatrix}
\end{align}

The following observations can be used to simplify the FN for an Assembler:
\begin{align}
&\frac{\partial \vec{P}_{ee}}{\partial \vec{P}_i}^T \frac{\partial \vec{P}_{ee}}{\partial \vec{P}_i} = \prod_{k=i-1}^{1}\mathbf{R}_k^T \prod_{k=1}^{i-1}\mathbf{R}_k = \mathbf{I} \\
&\frac{\partial \mathbf{R_{ee}}}{\partial \vec{P}_i}^T\frac{\partial \mathbf{R_{ee}}}{\partial \vec{P}_i} = \mathbf{0} \\
&\frac{\partial \mathbf{R_{ee}}}{\partial \theta_{i j}}^T\frac{\partial \mathbf{R_{ee}}}{\partial \theta_{i j}} = \prod_{k=1}^{i-1}(\mathbf{R_k^T})\frac{\partial \mathbf{R_i}}{\partial \theta_j}^T\prod_{k=i+1}^{n}\mathbf{R_k^T} \\ \notag
&\qquad \qquad \qquad \cdot \prod_{k=n}^{i+1}(\mathbf{R}_k)\frac{\partial \mathbf{R_i}}{\partial \theta_j}\prod_{k=i-1}^{1}\mathbf{R}_k = \mathbf{I}
\end{align}

\noindent
where the index $i$ denotes the platform number, and the index $j$ denotes the Euler angle axis for that platform's rotation. These equations show that large parts of $\mathbf{J^T_{ee}J_{ee}}$ are actually constant, meaning they need not be considered when minimizing the FN. Only the diagonal elements are necessary for computing the FN, so all the off diagonals can be ignored as well. This leaves only the submatrix $\frac{\partial \vec{P}_{ee}}{\partial \theta_{ij}}^T \frac{\partial \vec{P}_{ee}}{\partial \theta_{ij}}$, all of which are non-constant excepting $\frac{\partial \vec{P}_{ee}}{\partial \theta_{nj}} = 0$ where the $n^{th}$ index is the topmost platform. Also note that in~\eqreff{forward}, $\vec{P}_1$ does not have any factor dependent on the Euler angles. These facts mean that the FN does not depend on $\vec{P}_1$, nor on $\mathbf{R}_n$.

The inverse kinematic problem is further simplified by limiting the DOF of the Assembler to act within the plane containing the solution position. Let the desired $\vec{P}_{ee}$, which represents the end effector position in the reference frame of the base plate, be described in polar coordinates with $\rho$ as the radial, $\phi$ as the azimuthal angle, and $z$ as the z-axis coordinates. We hold $\phi$ constant and only work in the plane made by $\rho$ and $z$. We can then limit each platform's DOF to two translations and a rotation about the axis perpendicular to this plane. We therefore set $\vec{P}_i = [\rho_i, z_i]^T$ which reflects the 2D coordinates. The forward kinematics in 2D are then unchanged from \eqreff{forward}, except that rotation matrix products can now be simplified to:

\begin{align} \eqlabel{2Drote}
\prod_{i=1}^{n}\mathbf{R}_i =
\begin{bmatrix}
\cos(\sum_{i=1}^n \theta_i) & -\sin(\sum_{i=1}^n \theta_i) \\
\sin(\sum_{i=1}^n \theta_i) & \cos(\sum_{i=1}^n \theta_i)
\end{bmatrix}
\end{align}

We let $\phi$ define the axis of rotation $\hat{s}$ about which $\theta$ rotates, and we can convert from the 2D plane parameters into the 3D geometry. This is done with the following equations which describe $\hat{s}$, give a generic form of Rodrigues' rotation formula~\cite{belongie1999rodrigues} and translate from cylindrical coordinates to Cartesian:
\begin{align} \eqlabel{transfrom}
&\hat{s} =
\begin{bmatrix} \cos(\phi) \\ \sin(\phi) \\ 0 \end{bmatrix}  \times \begin{bmatrix} 0 \\ 0 \\ 1 \end{bmatrix} = \begin{bmatrix}
sin(\phi) \\ -cos(\phi) \\ 0
\end{bmatrix} \\
&\mathbf{R} = \cos(\theta) \mathbf{I} + \sin(\theta)[\hat{s}]_\times + (1-\cos(\theta)) \hat{s} \hat{s}^T \\
&\vec{P} = \begin{bmatrix}
\rho \cos(\phi) & \rho \sin(\phi) & z
\end{bmatrix}^T
\end{align}

Where $\theta$ is the rotation about $\hat{s}$, $[\hat{s}]_\times$ is the cross product matrix of $\hat{s}$, $\mathbf{R}$ is the rotation matrix about $\hat{s}$ by $\theta$, and $\vec{P}$ is the translation in Cartesian Coordinates. 

By simplifying the problem to two dimensional (2-D) space, we reduce half the DOF per platform and greatly simplify the orientation kinematics from matrix products to a simple sum. Because the axis of rotation is now determined by $\vec{P}_{ee}$, the orientation of the end effector is limited to one DOF.

\section{Optimization Problem}
The inverse kinematics of an Assembler are formulated as a constrained optimization problem. The goal is to minimize Trace($\mathbf{J_{ee}^T J_{ee}}$) while reaching the desired end effector position $\vec{P}_{ee}$ and angle $\theta_{ee}$. The optimization must also obey the constraints described in \eqreff{constraints} for each platform in the stack. MATLAB's Optimization Toolbox~\cite{MatlabOTB} is used to solve this optimization with the interior point method~\cite{ye1996interior}\cite{wright2005interior}\cite{potra2000interior}. This method works by iteratively approximating the full problem into sub-problems and solving them with linear approximation and trust-region solvers which approximate functions as quadratic~\cite{coleman1996interior}.

\section{Numerical Results}

Given an Assembler pose in the 2D plane defined by angle $\phi$, we can use Monte Carlo methods to simulate perturbations in the structure and test the resulting end effector movement. The variance of end effector poses is compared between an optimal pose and two suboptimal poses with the same desired end effector conditions. The suboptimal poses are found by running the same interior point solver, but without any minimization requirement so that the solver will return the first solution which meets the other constraints.

To generate an initial guess for the solver, a simple algorithm was designed which divides the translation and orientation among each platform while ignoring all other constraints. A minimum translation was imposed in each platform's $\hat{n}_b$ direction, so that the initial guess would be more reasonable. To find multiple solutions, we perturbed the initial guess by sampling from a normal distribution and re-running the solver until the desired number of successful runs are found.

\subsection{Procedure}
\algoref{perturb} details the method used to collect a Monte Carlo dataset. The function $ForwardKinematics()$ makes use of \eqreff{forward} and \eqreff{2Drote}.

\begin{algorithm}
	\caption{GetNPerturbations}
	\begin{algorithmic} \algolabel{perturb}
		\REQUIRE $N, pose_1, pose_2, pose_3, \phi, \sigma_t, \sigma_\theta$
		\FOR{n from 1 to N}
		\FORALL{Platforms i}
		\STATE $\delta x.\rho_i \leftarrow SampleGaussian(\sigma_t)$
		\STATE $\delta x.z_i \leftarrow SampleGaussian(\sigma_t)$
		\STATE $\delta x.\theta_i \leftarrow SampleGaussian(\sigma_\theta)$
		\ENDFOR
		\STATE $x_1 \leftarrow \delta x + pose_1$
		\STATE $x_2 \leftarrow \delta x + pose_2$
		\STATE $x_3 \leftarrow \delta x + pose_3$
		\STATE $y_{n1} \leftarrow ForwardKinematics(x_1, \phi)$
		\STATE $y_{n2} \leftarrow ForwardKinematics(x_2, \phi)$
		\STATE $y_{n3} \leftarrow ForwardKinematics(x_3, \phi)$
		\ENDFOR
	\end{algorithmic}
\end{algorithm}

A sample set of 10,000 data points was generated for a 4 platform Assembler. \figref{poses} shows an example of the Assembler poses that were examined. Each pose has the exact same end effector state, with $\rho = 600$~mm, $z = 1000$~mm, and $\theta = -1.57$~rad in the global frame. All perturbations were performed with $\sigma_\theta = 0.005$ rad, $\sigma_t = 1$ mm, unless otherwise stated.

\begin{figure}
	\includegraphics[width=\linewidth]{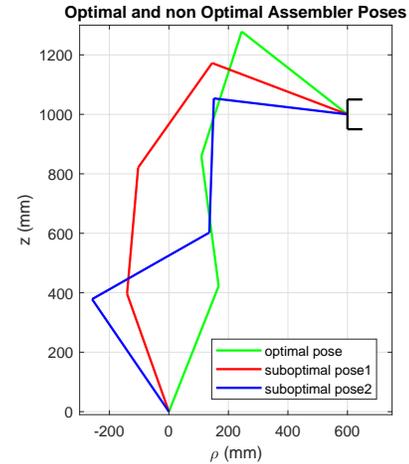}
	\caption{2D diagram of Assembler poses}
	\figlabel{poses}
\end{figure}

\subsection{Random Perturbation Results} 

\figref{perturbations1}-\figref{perturbations3} plot the perturbations from each sample taken with the poses in \figref{poses}. The red line indicates the 95\% confidence ellipse. The first line of \tabref{perturbData} shows the median distance the end effector moved, with the 95\% confidence interval. In this case the optimal pose led to significantly less variance than the others. The optimization led to approximately 15\% reduction in end effector movement.

\subsection{Linear Region Assumption}
Our optimization function assumes small perturbations in the region where a linear approximation of the forward kinematics is valid. We can test this assumption by estimating the covariance of end effector $\rho$ / $z$ positions and comparing to the Monte Carlo data.

A Gaussian sampling with a covariance matrix $\mathbf{C_x}$ undergoing a linear transformation given by matrix $\mathbf{J}$ will result in an equivalent Gaussian sampling with a covariance matrix given by 
\begin{align} \eqlabel{covariance}
\mathbf{C_y} = \mathbf{J C_x J^T}
\end{align}

In our case, the random perturbations of our state vector are described by a diagonal matrix of size 12 with values $\sigma_\theta^2$ and $\sigma_t^2$. We can compare the end effector pose observed covariance $\mathbf{C^{obs}}$ from the Monte Carlo perturbations from \algoref{perturb}, and the estimated covariance $\mathbf{C^{est}}$ from \eqreff{covariance} with:

\begin{align}
F = \frac{\Sigma_{i j} |\mathbf{C^{est}_{ij} - C^{obs}_{ij}|}}{\Sigma_{i j} |\mathbf{C^{obs}_{ij}}|}
\end{align}

where $\Sigma_{ij}$ represents a sum over all matrix indexes and $\mathbf{C_{i j}^*}$ is an element of $\mathbf{C^*}$. $F$ gives a proportional sum of differences between the two covariance matrices. A larger value of $F$ means the forward kinematics are less linear, a smaller value means the linear approximation is more accurate.

We set the end effector to reach 3 different states; for each desired end effector state, we found two suboptimal poses which did not apply any optimization and one optimal pose which minimized the FN. Sampling noise inputs were applied with standard deviations of $\sigma_\theta = .005$ rad and $\sigma_t = 1$ mm using 10,000 samples to \algoref{perturb}. This method was implemented for three different desired end effector conditions. The results are summarized in~\tabref{perturbData}. The values of $F$ for every pose at that noise level are $<1\%$, this value is small enough that there is significant variance across samplings. \tabref{linearData} gives the $F$ factor for much larger variance with 100,000 samples on the same poses as \tabref{perturbData}. The optimal poses consistently have a lower F-factor at high variance than the non-optimized poses. This observation of lower F values implies that it tends to take larger perturbations under optimized poses to break the linearity assumption than non-optimized poses.

\begin{figure}
	\includegraphics[width=\linewidth]{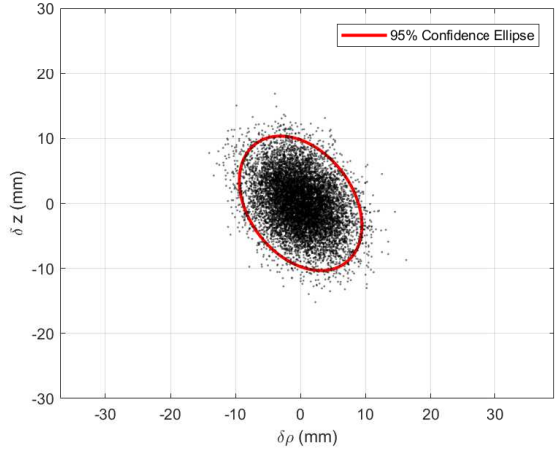}
	\caption{Optimal Pose End Effector Perturbations}
	\figlabel{perturbations1}
\end{figure}
\begin{figure}
	\includegraphics[width=\linewidth]{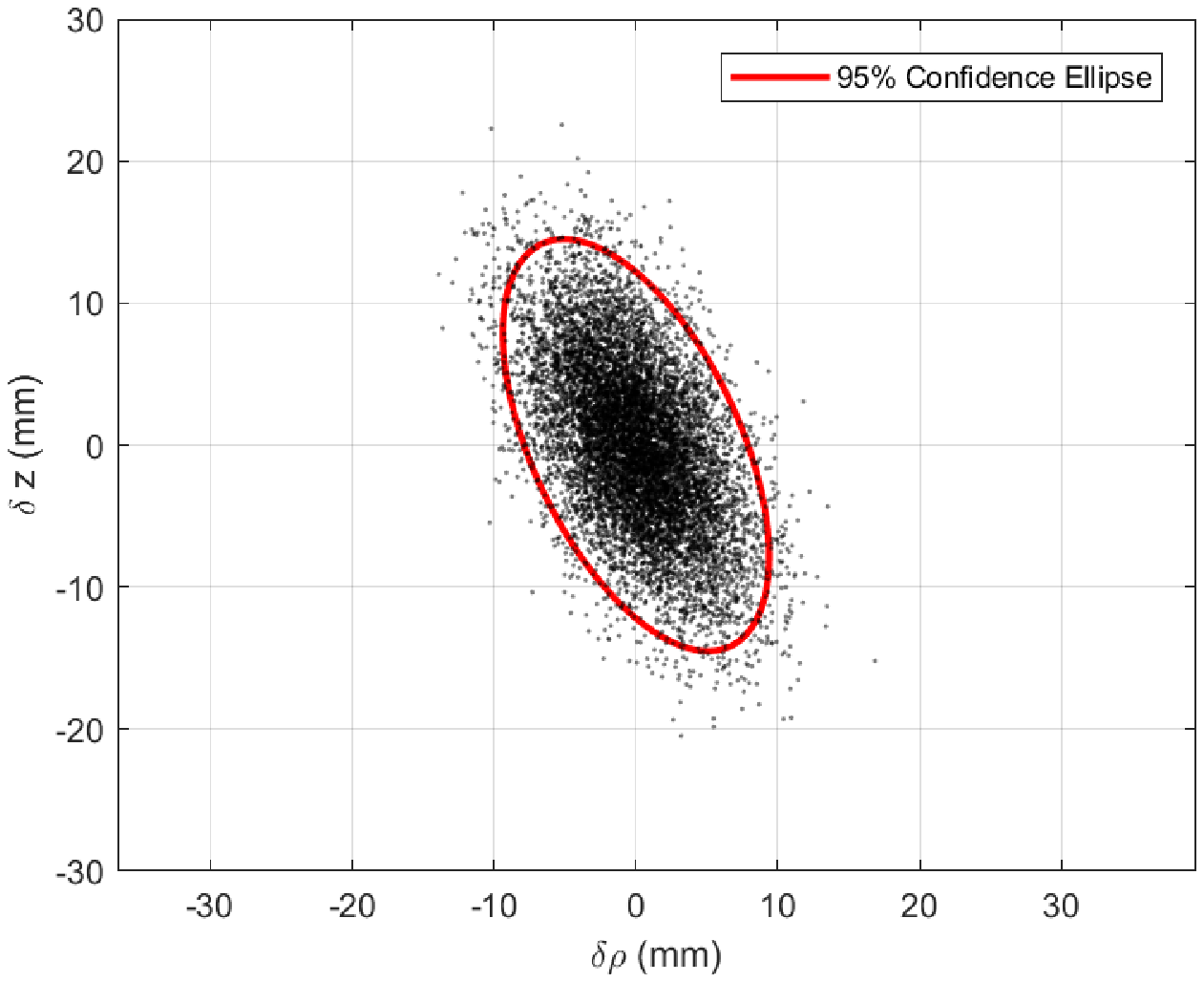}
	\caption{Suboptimal Pose1 End Effector Perturbations}
	\figlabel{perturbations2}
\end{figure}
\begin{figure}
	\includegraphics[width=\linewidth]{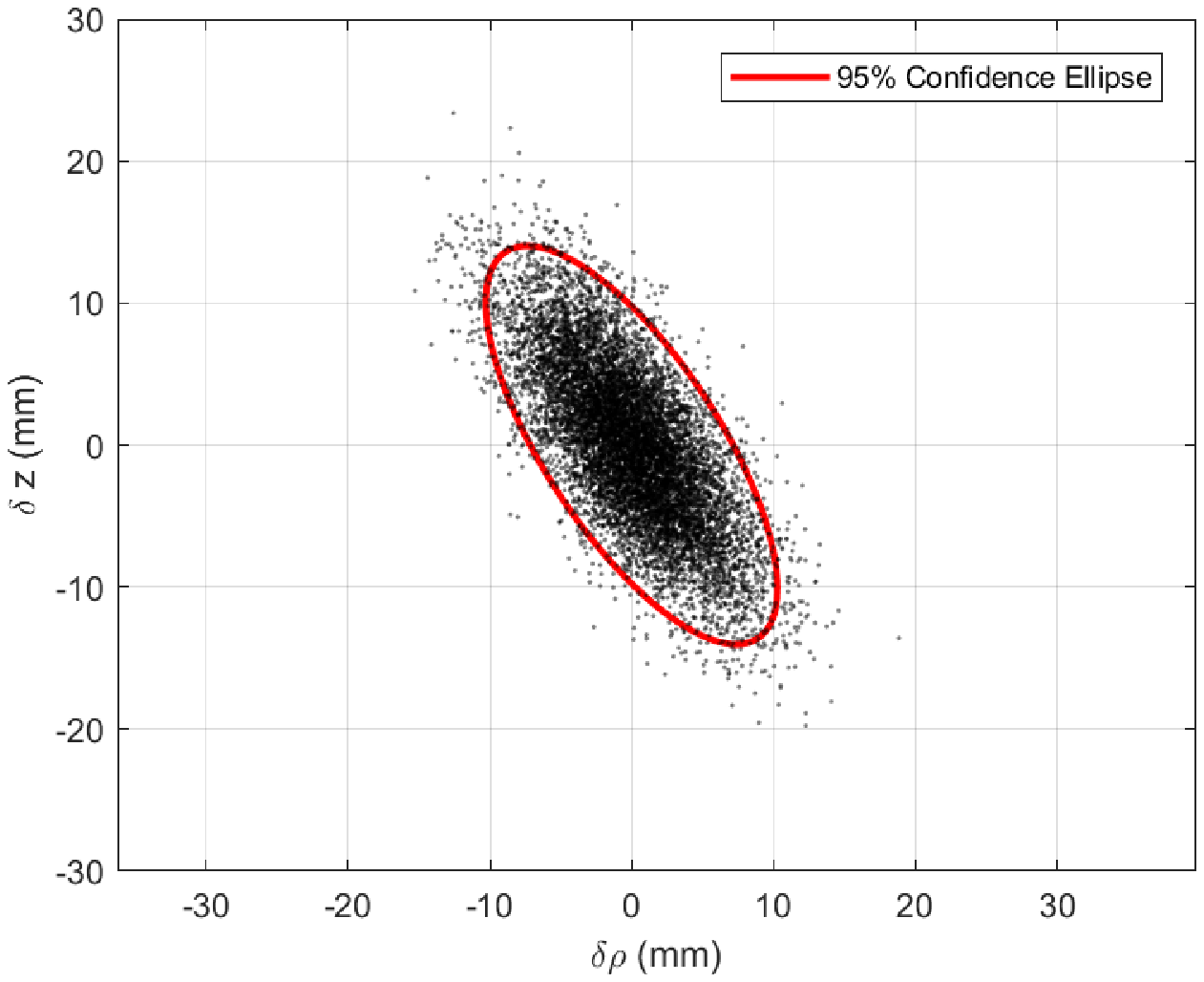}
	\caption{Suboptimal Pose2 End Effector Perturbations}
	\figlabel{perturbations3}
\end{figure}

\begin{table}
	\begin{tabular}{|c|c|c|c|}
		\hline
		Perturbation Data & Optimal Pose & Non-Opt 1 & Non-Opt 2 \\
		\hline
		$\vec{P}_{ee}=[600\ 1000]^T$ & 4.67 & 5.47 & 5.31 \\
		$\theta = -\frac{\pi}{2}$ rad & [0.9, 11.2] & [1.0, 14.7] & [1.0, 15.1] \\ \hline
		$\vec{P}_{ee}=[145\ 1500]^T$ & 5.10 & 5.40 & 5.56 \\
		$\theta = -0.207$ rad & [0.84, 15.3] & [0.87, 16.5] & [0.94, 16.8] \\ \hline
		$\vec{P}_{ee}=[-319\ 1532]^T$ & 5.32 & 5.58 & 5.69 \\
		$\theta = 0.332$ rad & [0.89, 16.4] & [0.93, 17.4] & [0.92, 17.8] \\ \hline
	\end{tabular}
	\caption{Median distance in mm moved by end effector at different noise levels with 95\% confidence interval in brackets; $\vec{P}_{ee}$ given in mm}
	\tablabel{perturbData}
\end{table}

\begin{table}
	\begin{tabular}{|c|c|c|c|}
		\hline
		F factor at $\sigma$ & Optimal Pose & Non-Opt 1 & Non-Opt 2  \\
		\hline
		$\vec{P}_{ee}=[600\ 1000]^T$ & 25\% & 47\% & 45\% \\
		\hline
		$\vec{P}_{ee}=[145\ 1500]^T$ & 63\% & 64\% & 66\% \\
		\hline
		$\vec{P}_{ee}=[-319\ 1532]^T$ & 61\% & 64\% & 70\% \\
		\hline
	\end{tabular}
	\caption{F-factor values for each pose with high variance $\sigma_\theta = 0.5$ rad; $\sigma_t = 100$ mm; $\phi = 0$ rad; $\vec{P}_{ee}$ given in mm}
	\tablabel{linearData}
\end{table}

\figref{ratio} shows a plot of all found solutions colored by the FN ratio between an optimized pose and a non-optimized pose. The lower value colors indicate the optimal pose was more successful. These points tend to be concentrated in the center of the distribution because the DOFs are less constrained and the optimizer has more feasible poses to choose from.

\begin{figure}
	\includegraphics[width=\linewidth]{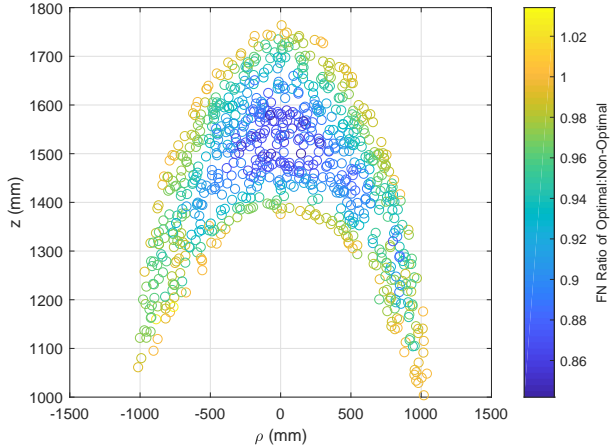}
	\caption{Plot of Found Solutions colored by FN ratio between optimal and non-optimal poses}
	\figlabel{ratio}
\end{figure}

\figref{MedianRatio} empirically shows the relationship between the FN and the perturbation distance. A simple linear regression shows an $r^2$ value of 0.97, which demonstrates a strong linear correlation. There is a linear slope of 0.4, this means that a reduction in the FN by 1\% is expected to cause a reduction in perturbation distance by 0.4\%.

\begin{figure}
	\includegraphics[width=\linewidth]{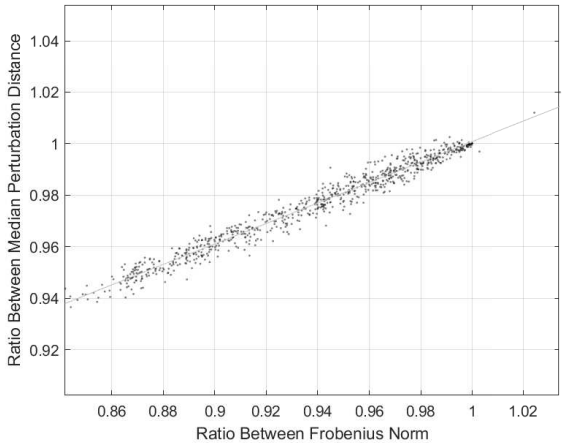}
	\caption{Plot of ratio between optimized and non-optimized poses of FN value and Median Perturbation distance}
	\figlabel{MedianRatio}
\end{figure}

\section{Concluding Remarks}
The Frobenius Norm methodology was used to optimize the pose selection of over-actuated structures with many degrees of freedom and non-linear forward kinematics. This method was evaluated with a Monte Carlo simulation of Assemblers. Poses with optimized Frobenius Norms consistently outperformed the non-optimized poses due to the geometric properties of the linear approximation used. This method can be applied to any over-actuated structure for which the forward kinematics are known using conventional methods. Demonstration of Frobenius Norm minimization on other geometries with hardware validation is proposed for future efforts.


\bibliographystyle{ACM-Reference-Format}  
\bibliography{ref}  


\begin{thebibliography}{25}


\ifx \showCODEN    \undefined \def \showCODEN     #1{\unskip}     \fi
\ifx \showDOI      \undefined \def \showDOI       #1{#1}\fi
\ifx \showISBNx    \undefined \def \showISBNx     #1{\unskip}     \fi
\ifx \showISBNxiii \undefined \def \showISBNxiii  #1{\unskip}     \fi
\ifx \showISSN     \undefined \def \showISSN      #1{\unskip}     \fi
\ifx \showLCCN     \undefined \def \showLCCN      #1{\unskip}     \fi
\ifx \shownote     \undefined \def \shownote      #1{#1}          \fi
\ifx \showarticletitle \undefined \def \showarticletitle #1{#1}   \fi
\ifx \showURL      \undefined \def \showURL       {\relax}        \fi
\providecommand\bibfield[2]{#2}
\providecommand\bibinfo[2]{#2}
\providecommand\natexlab[1]{#1}
\providecommand\showeprint[2][]{arXiv:#2}

\bibitem[\protect\citeauthoryear{Abdellatif and Heimann}{Abdellatif and
  Heimann}{2009}]%
        {abdellatif2009computational}
\bibfield{author}{\bibinfo{person}{Houssem Abdellatif} {and}
  \bibinfo{person}{Bodo Heimann}.} \bibinfo{year}{2009}\natexlab{}.
\newblock \showarticletitle{Computational efficient inverse dynamics of 6-DOF
  fully parallel manipulators by using the Lagrangian formalism}.
\newblock \bibinfo{journal}{\emph{Mechanism and Machine Theory}}
  \bibinfo{volume}{44}, \bibinfo{number}{1} (\bibinfo{year}{2009}),
  \bibinfo{pages}{192--207}.
\newblock


\bibitem[\protect\citeauthoryear{Belongie}{Belongie}{1999}]%
        {belongie1999rodrigues}
\bibfield{author}{\bibinfo{person}{Serge Belongie}.}
  \bibinfo{year}{1999}\natexlab{}.
\newblock \showarticletitle{Rodrigues’ rotation formula}.
\newblock \bibinfo{journal}{\emph{From MathWorld--A Wolfram Web Resource,
  created by Eric W. Weisstein. http://mathworld. wolfram.
  com/RodriguesRotationFormula. html}} (\bibinfo{year}{1999}).
\newblock


\bibitem[\protect\citeauthoryear{Bingul and Karahan}{Bingul and
  Karahan}{2012}]%
        {bingul2012dynamic}
\bibfield{author}{\bibinfo{person}{Zafer Bingul} {and} \bibinfo{person}{Oguzhan
  Karahan}.} \bibinfo{year}{2012}\natexlab{}.
\newblock \showarticletitle{Dynamic modeling and simulation of Stewart
  platform}.
\newblock In \bibinfo{booktitle}{\emph{Serial and Parallel Robot
  Manipulators-Kinematics, Dynamics, Control and Optimization}}.
  \bibinfo{publisher}{InTech}, \bibinfo{pages}{19--42}.
\newblock


\bibitem[\protect\citeauthoryear{Cohen and Kennedy}{Cohen and Kennedy}{1997}]%
        {cohen1997habitats}
\bibfield{author}{\bibinfo{person}{Marc Cohen} {and} \bibinfo{person}{Kriss~J
  Kennedy}.} \bibinfo{year}{1997}\natexlab{}.
\newblock \showarticletitle{Habitats and surface construction: technology and
  development roadmap}. In \bibinfo{booktitle}{\emph{NASA conference
  publication}}, Vol.~\bibinfo{volume}{CP-97-206241}. NASA,
  \bibinfo{pages}{75--96}.
\newblock


\bibitem[\protect\citeauthoryear{Coleman and Li}{Coleman and Li}{1996}]%
        {coleman1996interior}
\bibfield{author}{\bibinfo{person}{Thomas~F Coleman} {and}
  \bibinfo{person}{Yuying Li}.} \bibinfo{year}{1996}\natexlab{}.
\newblock \showarticletitle{An interior trust region approach for nonlinear
  minimization subject to bounds}.
\newblock \bibinfo{journal}{\emph{SIAM Journal on optimization}}
  \bibinfo{volume}{6}, \bibinfo{number}{2} (\bibinfo{year}{1996}),
  \bibinfo{pages}{418--445}.
\newblock


\bibitem[\protect\citeauthoryear{Cust{\'o}dio, Rocha, and Vicente}{Cust{\'o}dio
  et~al\mbox{.}}{2010}]%
        {custodio2010incorporating}
\bibfield{author}{\bibinfo{person}{Ana~Lu{\'\i}sa Cust{\'o}dio},
  \bibinfo{person}{Humberto Rocha}, {and} \bibinfo{person}{Lu{\'\i}s~N
  Vicente}.} \bibinfo{year}{2010}\natexlab{}.
\newblock \showarticletitle{Incorporating minimum Frobenius norm models in
  direct search}.
\newblock \bibinfo{journal}{\emph{Computational Optimization and Applications}}
  \bibinfo{volume}{46}, \bibinfo{number}{2} (\bibinfo{year}{2010}),
  \bibinfo{pages}{265--278}.
\newblock


\bibitem[\protect\citeauthoryear{Doggett}{Doggett}{2002}]%
        {doggett2002robotic}
\bibfield{author}{\bibinfo{person}{William Doggett}.}
  \bibinfo{year}{2002}\natexlab{}.
\newblock \showarticletitle{Robotic assembly of truss structures for space
  systems and future research plans}. In \bibinfo{booktitle}{\emph{Aerospace
  Conference Proceedings, 2002. IEEE}}, Vol.~\bibinfo{volume}{7}. IEEE,
  \bibinfo{pages}{7--7}.
\newblock


\bibitem[\protect\citeauthoryear{Hamlin and Sanderson}{Hamlin and
  Sanderson}{1996}]%
        {hamlin1996tetrobot}
\bibfield{author}{\bibinfo{person}{Gregory~J Hamlin} {and}
  \bibinfo{person}{Arthur~C Sanderson}.} \bibinfo{year}{1996}\natexlab{}.
\newblock \showarticletitle{Tetrobot modular robotics: Prototype and
  experiments}. In \bibinfo{booktitle}{\emph{Intelligent Robots and Systems'
  96, IROS 96, Proceedings of the 1996 IEEE/RSJ International Conference on}},
  Vol.~\bibinfo{volume}{2}. IEEE, \bibinfo{pages}{390--395}.
\newblock


\bibitem[\protect\citeauthoryear{Howe, Kennedy, Gill, Smith, and George}{Howe
  et~al\mbox{.}}{2013}]%
        {howe2013nasa}
\bibfield{author}{\bibinfo{person}{Scott~A Howe}, \bibinfo{person}{Kriss~J
  Kennedy}, \bibinfo{person}{Tracy~R Gill}, \bibinfo{person}{Russell~W Smith},
  {and} \bibinfo{person}{Patrick George}.} \bibinfo{year}{2013}\natexlab{}.
\newblock \showarticletitle{NASA habitat demonstration unit (HDU) deep space
  habitat analog}. In \bibinfo{booktitle}{\emph{AIAA SPACE 2013 Conference and
  Exposition}}. \bibinfo{pages}{5436}.
\newblock


\bibitem[\protect\citeauthoryear{Jakobovi{\'c} and Budin}{Jakobovi{\'c} and
  Budin}{2002}]%
        {jakobovic2002forward}
\bibfield{author}{\bibinfo{person}{Domagoj Jakobovi{\'c}} {and}
  \bibinfo{person}{Leo Budin}.} \bibinfo{year}{2002}\natexlab{}.
\newblock \showarticletitle{Forward kinematics of a Stewart platform
  mechanism}.
\newblock \bibinfo{journal}{\emph{Faculty of Electrical Engineering and
  Computing, Unska, Zagreb, Croatia}} (\bibinfo{year}{2002}).
\newblock


\bibitem[\protect\citeauthoryear{Kennedy}{Kennedy}{2011}]%
        {kennedy2011nasa}
\bibfield{author}{\bibinfo{person}{Kriss~J Kennedy}.}
  \bibinfo{year}{2011}\natexlab{}.
\newblock \showarticletitle{NASA Habitat Demonstration Unit Project--Deep Space
  Habitat Overview}. In \bibinfo{booktitle}{\emph{41st International Conference
  on Environmental Systems (ICES), Portland, Oregon, USA}}.
  \bibinfo{pages}{17--21}.
\newblock


\bibitem[\protect\citeauthoryear{Komendera, Dorsey, Doggett, and
  Correll}{Komendera et~al\mbox{.}}{2014}]%
        {komendera2014truss}
\bibfield{author}{\bibinfo{person}{E. Komendera}, \bibinfo{person}{J.~T.
  Dorsey}, \bibinfo{person}{W.~R. Doggett}, {and} \bibinfo{person}{N.
  Correll}.} \bibinfo{year}{2014}\natexlab{}.
\newblock \showarticletitle{Truss assembly and welding by Intelligent Precision
  Jigging Robots}. In \bibinfo{booktitle}{\emph{2014 IEEE International
  Conference on Technologies for Practical Robot Applications (TePRA)}}.
  \bibinfo{pages}{1--6}.
\newblock
\showISSN{2325-0534}
\urldef\tempurl%
\url{https://doi.org/10.1109/TePRA.2014.6869150}
\showDOI{\tempurl}


\bibitem[\protect\citeauthoryear{MATLAB Optimization Toolbox}{MATLAB
  Optimization Toolbox}{2017}]%
        {MatlabOTB}
MATLAB Optimization Toolbox \bibinfo{year}{2017}\natexlab{}.
\newblock \bibinfo{title}{MATLAB Optimization Toolbox}.
\newblock
\newblock
\newblock
\shownote{The MathWorks, Natick, MA, USA.}


\bibitem[\protect\citeauthoryear{Mikulas~Jr, Rhodes, and Simonton}{Mikulas~Jr
  et~al\mbox{.}}{1987}]%
        {mikulas1987deployable}
\bibfield{author}{\bibinfo{person}{Martin~M Mikulas~Jr},
  \bibinfo{person}{Marvin~D Rhodes}, {and} \bibinfo{person}{J~Wayne Simonton}.}
  \bibinfo{year}{1987}\natexlab{}.
\newblock \bibinfo{title}{Deployable geodesic truss structure}.
\newblock
\newblock
\newblock
\shownote{US Patent 4,677,803.}


\bibitem[\protect\citeauthoryear{Miura and Furuya}{Miura and Furuya}{1988}]%
        {miura1988adaptive}
\bibfield{author}{\bibinfo{person}{Koryo Miura} {and} \bibinfo{person}{Hiroshi
  Furuya}.} \bibinfo{year}{1988}\natexlab{}.
\newblock \showarticletitle{Adaptive structure concept for future space
  applications}.
\newblock \bibinfo{journal}{\emph{AIAA journal}} \bibinfo{volume}{26},
  \bibinfo{number}{8} (\bibinfo{year}{1988}), \bibinfo{pages}{995--1002}.
\newblock


\bibitem[\protect\citeauthoryear{Potra and Wright}{Potra and Wright}{2000}]%
        {potra2000interior}
\bibfield{author}{\bibinfo{person}{Florian~A Potra} {and}
  \bibinfo{person}{Stephen~J Wright}.} \bibinfo{year}{2000}\natexlab{}.
\newblock \showarticletitle{Interior-point methods}.
\newblock \bibinfo{journal}{\emph{J. Comput. Appl. Math.}}
  \bibinfo{volume}{124}, \bibinfo{number}{1-2} (\bibinfo{year}{2000}),
  \bibinfo{pages}{281--302}.
\newblock


\bibitem[\protect\citeauthoryear{Rhodes and Mikulas~Jr}{Rhodes and
  Mikulas~Jr}{1985}]%
        {rhodes1985deployable}
\bibfield{author}{\bibinfo{person}{Marvin~D Rhodes} {and} \bibinfo{person}{MM
  Mikulas~Jr}.} \bibinfo{year}{1985}\natexlab{}.
\newblock \showarticletitle{Deployable controllable geometry truss beam}.
\newblock \bibinfo{journal}{\emph{NASA Langley Research Center Report
  TM-86366}} (\bibinfo{year}{1985}).
\newblock


\bibitem[\protect\citeauthoryear{Roa, Nottensteiner, Wedler, and Grunwald}{Roa
  et~al\mbox{.}}{2017}]%
        {roarobotic}
\bibfield{author}{\bibinfo{person}{M{\'a}ximo~A Roa},
  \bibinfo{person}{Korbinian Nottensteiner}, \bibinfo{person}{Armin Wedler},
  {and} \bibinfo{person}{Gerhard Grunwald}.} \bibinfo{year}{2017}\natexlab{}.
\newblock \showarticletitle{Robotic Technologies for In-Space Assembly
  Operations}.
\newblock \bibinfo{journal}{\emph{Symposium on Advanced Space Technologies in
  Robotics and Automation}} (\bibinfo{year}{2017}).
\newblock


\bibitem[\protect\citeauthoryear{Rojas and Peters}{Rojas and Peters}{2012}]%
        {rojas2012analysis}
\bibfield{author}{\bibinfo{person}{Juan Rojas} {and} \bibinfo{person}{Richard~A
  Peters}.} \bibinfo{year}{2012}\natexlab{}.
\newblock \showarticletitle{Analysis of autonomous cooperative assembly using
  coordination schemes by heterogeneous robots using a control basis approach}.
\newblock \bibinfo{journal}{\emph{Autonomous Robots}} \bibinfo{volume}{32},
  \bibinfo{number}{4} (\bibinfo{year}{2012}), \bibinfo{pages}{369--383}.
\newblock


\bibitem[\protect\citeauthoryear{Watson, Collins, and Bush}{Watson
  et~al\mbox{.}}{2002}]%
        {watson2002history}
\bibfield{author}{\bibinfo{person}{Judith~J Watson}, \bibinfo{person}{Timothy~J
  Collins}, {and} \bibinfo{person}{Harold~G Bush}.}
  \bibinfo{year}{2002}\natexlab{}.
\newblock \showarticletitle{A history of astronaut construction of large space
  structures at NASA Langley Research Center}. In
  \bibinfo{booktitle}{\emph{Aerospace Conference Proceedings, 2002. IEEE}},
  Vol.~\bibinfo{volume}{7}. IEEE, \bibinfo{pages}{7--7}.
\newblock


\bibitem[\protect\citeauthoryear{Williams et~al\mbox{.}}{Williams
  et~al\mbox{.}}{1994}]%
        {williams1994kinematic}
\bibfield{author}{\bibinfo{person}{Robert~L Williams} {et~al\mbox{.}}}
  \bibinfo{year}{1994}\natexlab{}.
\newblock \showarticletitle{Kinematic modeling of a double octahedral Variable
  Geometry Truss (VGT) as an extensible gimbal}.
\newblock \bibinfo{journal}{\emph{NASA Technical Memorandum 109127}}
  (\bibinfo{year}{1994}).
\newblock


\bibitem[\protect\citeauthoryear{Wright}{Wright}{2005}]%
        {wright2005interior}
\bibfield{author}{\bibinfo{person}{Margaret Wright}.}
  \bibinfo{year}{2005}\natexlab{}.
\newblock \showarticletitle{The interior-point revolution in optimization:
  history, recent developments, and lasting consequences}.
\newblock \bibinfo{journal}{\emph{Bulletin of the American mathematical
  society}} \bibinfo{volume}{42}, \bibinfo{number}{1} (\bibinfo{year}{2005}),
  \bibinfo{pages}{39--56}.
\newblock


\bibitem[\protect\citeauthoryear{Ye}{Ye}{1996}]%
        {ye1996interior}
\bibfield{author}{\bibinfo{person}{Yinyu Ye}.} \bibinfo{year}{1996}\natexlab{}.
\newblock \showarticletitle{Interior-Point Algorithm: Theory and Practice}.
\newblock  (\bibinfo{year}{1996}).
\newblock


\bibitem[\protect\citeauthoryear{Yim, Zhang, and Duff}{Yim
  et~al\mbox{.}}{2002}]%
        {yim2002modular}
\bibfield{author}{\bibinfo{person}{Mark Yim}, \bibinfo{person}{Ying Zhang},
  {and} \bibinfo{person}{David Duff}.} \bibinfo{year}{2002}\natexlab{}.
\newblock \showarticletitle{Modular robots}.
\newblock \bibinfo{journal}{\emph{IEEE Spectrum}} \bibinfo{volume}{39},
  \bibinfo{number}{2} (\bibinfo{year}{2002}), \bibinfo{pages}{30--34}.
\newblock


\bibitem[\protect\citeauthoryear{Zimpfer, Kachmar, and Tuohy}{Zimpfer
  et~al\mbox{.}}{2005}]%
        {zimpfer2005autonomous}
\bibfield{author}{\bibinfo{person}{Douglas Zimpfer}, \bibinfo{person}{Peter
  Kachmar}, {and} \bibinfo{person}{Seamus Tuohy}.}
  \bibinfo{year}{2005}\natexlab{}.
\newblock \showarticletitle{Autonomous rendezvous, capture and in-space
  assembly: past, present and future}. In \bibinfo{booktitle}{\emph{1st Space
  exploration conference: continuing the voyage of discovery}}.
  \bibinfo{pages}{2523}.
\newblock


\end{thebibliography}

\end{document}